\documentclass{ecai} 

\usepackage[utf8]{inputenc}

\usepackage{latexsym}
\usepackage{amssymb}
\usepackage{amsmath}
\usepackage{amsthm}
\usepackage{booktabs}
\usepackage{enumitem}
\usepackage{graphicx}
\usepackage{color}

\newcommand{\BibTeX}{B\kern-.05em{\sc i\kern-.025em b}\kern-.08em\TeX}

\usepackage{float}
\begin{document}


\begin{frontmatter}




\title{Replay to Remember (R2R): An Efficient Uncertainty-driven Unsupervised Continual Learning Framework Using Generative Replay}


\author[1]{\fnms{Sriram}~\snm{Mandalika}\orcid{0009-0009-8390-274X}}
\author[1]{\fnms{Harsha}~\snm{Vardhan}}
\author[1]{\fnms{Athira}~\snm{Nambiar}\orcid{0000-0002-4957-5804}\thanks{Corresponding Author. Email: athiram@srmist.edu.in}} 

\address[1]{Department of Computational Intelligence, SRM Institute of Science and Technology, Chennai, India}


\begin{abstract}
Continual Learning entails progressively acquiring knowledge from new data while retaining previously acquired knowledge, thereby mitigating ``Catastrophic Forgetting" in neural networks. Our work presents a novel uncertainty-driven Unsupervised Continual Learning framework using Generative Replay, namely \textbf{``Replay to Remember (R2R)''}. The proposed R2R architecture efficiently uses unlabelled and synthetic labelled data in a balanced proportion using a cluster-level uncertainty-driven feedback mechanism and a VLM-powered generative replay module. \textcolor{black}{Unlike traditional memory-buffer methods that depend on pretrained models and pseudo-labels, our R2R framework operates without any prior training. It leverages visual features from unlabeled data and adapts continuously using clustering-based uncertainty estimation coupled with dynamic thresholding.} \textcolor{black}{Concurrently, a generative replay mechanism along with DeepSeek-R1 powered CLIP VLM produces labelled synthetic data representative of past experiences, resembling biological visual thinking that \textbf{\textit{replays}} memory to \textbf{\textit{remember and act}} in new unseen tasks. }Extensive experimental analyses are carried out in CIFAR-10, CIFAR-100, CINIC-10, SVHN and TinyImageNet datasets. \textcolor{black}{Our proposed R2R approach improves knowledge retention, achieving a state-of-the-art performance of 98.13\%, 73.06\%, 93.41\%, 95.18\%, 59.74\% respectively, surpassing state-of-the-art performance by over 4.36\%}. 
\end{abstract}

\end{frontmatter}

\vspace{-9pt}
\section{Introduction}

Traditional machine learning assumes access to independently and identically distributed data in bulk, allowing models to generalize effectively from representative samples. This leads to \textit{'catastrophic forgetting'} of the data is previously learnt \cite{vandeVen2024ContinualLA}. To address this, Continual Learning (CL), which enables the gradual learning of new information while retaining knowledge gained from earlier data, has emerged as a solution \cite{10190202}. However, CL still faces challenges, notably the persistent issue of catastrophic forgetting. To mitigate this, various strategies have emerged, including including Elastic Weight Consolidation (EWC) \cite{Kirkpatrick2017OvercomingCF} which constrains updates to crucial weights; Experience Replay (ER) \cite{Rolnick2019ExperienceRF}, that stores and reuses a buffer of past examples; and Progressive Neural Networks (PNNs) \cite{Rolnick2019ExperienceRF}, which dedicate separate subnetworks for each new task.

Despite the effectiveness of these methods, many traditional CL approaches rely on CNN-based architectures pretrained on large-scale datasets like ImageNet \cite{Russakovsky2014ImageNetLS}. While effective in supervised settings, these methods assume \textit{labelled data or explicit task boundaries}, limiting their applicability in real-world scenarios where such supervision is unavailable. While pretrained weights offer useful priors, \textit{they lack adaptability to non-stationary distributions}, exacerbating the stability-plasticity dilemma \cite{Kim2023OnTS} and leading to catastrophic forgetting. Moreover, these models often \textit{fail to capture uncertainty} in latent spaces, particularly in regions with vague or overlapping class structures. This results in weak clustering and poor generalization. Static architectural constraints further reduce their flexibility to model complexity or evolving semantics without costly retraining \cite{Ahn2019UncertaintybasedCL, Zhou2023ClassIncrementalLA, Yin2024AnER, Liu2020AdaptiveAN, Zhou2023RevisitingCL}.


\textcolor{black}{To address these challenges, recent works have explored \textbf{Generative Replay} \cite{Khan2023LookingTT, Shin2017ContinualLW}, as an alternative to buffer-based methods. In particular, Class Incremental Learning (CIL) \cite{Masana2020ClassIncrementalLS} settings have shown that regularization alone, without access to exemplars, yields limited performance. Generative models like GANs \cite{Goodfellow2014GenerativeAN} and diffusion-based architectures \cite{Podell2023SDXLIL} offer a promising avenue for memory-free replay by synthesizing representative samples from earlier tasks. However, most of the existing CL methods apply generative replay uniformly \cite{Khan2024BrainInspiredCL, Shin2017ContinualLW}, ignoring the varying reliability of learned representations. To address this limitation, we propose a novel unsupervised continual learning framework, \textit{viz.} \textbf{``Replay to Remember (R2R)"}, which enhances knowledge retention by selectively reinforcing weak clusters through semantically guided generative replay, enabling effective adaptation without labeled data. \textit{To the best of our knowledge, R2R marks one of the first unsupervised CL frameworks using generative replay}. }



In detail, R2R framework presents a novel statistically oriented uncertainty estimation and VLM-powered generative replay for Continual learning (CL) setting. In particular, it consists of an initial \textbf{Frontier stage} that uses a Convolutional Autoencoder as a main model to classify unlabelled samples. This classification is performed using the Gaussian Mixture Model (GMM) for clusters using the latent vectors from the encoder part of CAE. Further, in the next stage, a newly proposed \textbf{Self-Guided Uncertainty-driven Feedback Mechanism (SG-UDFM)} identifies ambiguous clusters via. a novel ``statistically oriented thresholding mechanism" to extract the relevant samples to \textit{`replay and remember'}. In the next stage, a \textbf{VLM-powered generative replay} module utilizes a state-of-the-art diffusion model work from \cite{Rombach2021HighResolutionIS} for generating synthetic data as part of the VLM-powered Generative Replay (GR) module. In particular,  VLM-powered by DeepSeek-R1 \cite{DeepSeekAI2025DeepSeekR1IR} to Contrastive Language-Image Pretraining (CLIP) is utilized for mapping the words to the generated samples from the diffusion model. This mimics the biological visual thinking capability in intelligent beings to act in unseen tasks by replaying and remembering memories from past and adapting to the current relevant task \cite{Ewell2014ReplayTR}. Both of the aforesaid modules, `SG-UDFM' and `VLM-powered generative replay', play a vital role in training the framework in the absence of labelled data. At the final \textbf{self-improvement} stage, cluster-wise fine-tuning is facilitated to improve the overall system performance. 

Extensive qualitative and quantitative analysis of the R2R continual learning framework is carried out in the datasets CIFAR-10, CIFAR-100, CINIC-10, SVHN-10 and TinyImageNet. The R2R framework mitigates catastrophic forgetting issues and minimises the discrepancy between original and generated features for image classification, improving final model accuracy over multiple subsequent tasks. The key contributions of the paper are:
\vspace{-.2cm}
\begin{itemize}[noitemsep]
    \item Proposal of a novel visual thinking-based Continual Learning setting, namely \textbf{``Replay to Remember' (R2R)- An  Uncertainty-driven Unsupervised Continual Learning Framework Using Generative Replay} for image classification, first-of-its-kind.
    \item Proposal of novel statistically oriented \textbf{self-guided uncertainty-driven feedback mechanism (SG-UDFM)} via. clustering-based uncertainty estimation and dynamic thresholding.
    \item Use of \textbf{No Pretained Policy}, a significant achievement against conventional state-of-the-art approaches.  
\end{itemize}
\vspace{-.2cm}

The rest of the paper is organized as follows. The related works are described in Section 2. The proposed `Replay to Remember (R2R)' framework is presented in Section 3. In Section 4 and Section 5, the experimental setup and the results are discussed in detail. Finally, the summary of the paper and some future plans are enumerated in Section 6.

\vspace{-.3cm}
\section{Related Works}

Continual learning has been a longstanding challenge in artificial intelligence, particularly in mitigating catastrophic forgetting while ensuring efficient knowledge retention. Several approaches have been proposed in recent years to address these challenges, including memory-based techniques, knowledge distillation, and generative replay.

\vspace{-11pt}
\subsection{Memory-Based Approaches}
Memory replay techniques store past experiences in an episodic memory buffer to enable selective rehearsal during learning. Methods such as Incremental Classifier and Representation Learning (iCaRL) \cite{Rebuffi2016iCaRLIC} employ exemplar sets to represent old classes and perform nearest-neighbor classification to balance past and current knowledge. However, these approaches suffer from storage constraints and privacy concerns due to the need to retain original data \cite{design_radhakrishnan_2024}. Another approach, Gradient Episodic Memory (GEM), prevents task interference by projecting gradients onto an allowable subspace. While effective in reducing forgetting, its dependence on stored past data limits scalability\cite{Liu2024RethinkingKD}.


\vspace{-11pt}
\subsection{Knowledge Distillation for Continual Learning}
Knowledge distillation methods transfer information from a teacher to a student model to improve continual learning. One approach leverages noisy student self-training, where a pretrained model generates adaptive pseudo-labels for new data, ensuring effective knowledge transfer \cite{Kumari2023ContinualLI}. Similarly, a novel knowledge distillation technique for semantic segmentation minimizes feature discrepancies between teacher and student models using angular-based loss functions \cite{fu2024unsupervised}. While effective, these methods require precise hyperparameter tuning, which can be challenging across diverse datasets \cite{Goswami2024ResurrectingOC}.

\vspace{-11pt}
\subsection{Generative Replay for Knowledge Retention}
Deep Generative Replay employs a generative model to replay synthetic samples of past tasks. Recent advancements include diffusion-based generative replay, where stable diffusion models generate high-fidelity synthetic samples to improve class-incremental learning \cite{Khan2023LookingTT}. However, maintaining the quality of synthetic data remains a challenge \cite{Tang2023KaizenPS}. A novel approach leverages a latent matching loss to align latent representations of generated samples with real data, ensuring improved retention and minimizing reconstruction misalignment \cite{Khan2024BrainInspiredCL}. Additionally, another method proposed an adversarial approach to perturb new data into past class prototypes, allowing for more effective class recovery without explicit storage of previous samples \cite{Jodelet2023ClassIncrementalLU}.

\vspace{-11pt}
\subsection{Uncertainty-driven Continual Learning}
Recent works have explored uncertainty-driven mechanisms to improve continual learning. For instance, entropy-based uncertainty measures identify weak class representations and reinforce them using generative models in \cite{Shin2017ContinualLW}. This aligns with works that introduced rememory-based techniques to retain past knowledge without relying on explicit memory buffers \cite{Krukowski2024HyperIntervalHA}.

\textcolor{black}{In contrast to prior works that predominantly focus on memory buffer-based techniques for storing past samples, our R2R leverages \textbf{Generative Replay (GR)} to synthesize data dynamically, eliminating the need for explicit storage. Furthermore, it employs an \textbf{uncertainty-driven} mechanism to refine high-uncertainty clusters, ensuring efficient and targeted adaptation selectively. This enables \textbf{self-guided learning}, where synthetic samples are generated on demand, preserving task-relevant knowledge while mitigating catastrophic forgetting.}

\vspace{-11pt}
\section{Methodology: Replay to Remember (R2R)}

\begin{figure*}[t]
    \centering
    \includegraphics[width=0.85\textwidth, height=8cm, keepaspectratio]{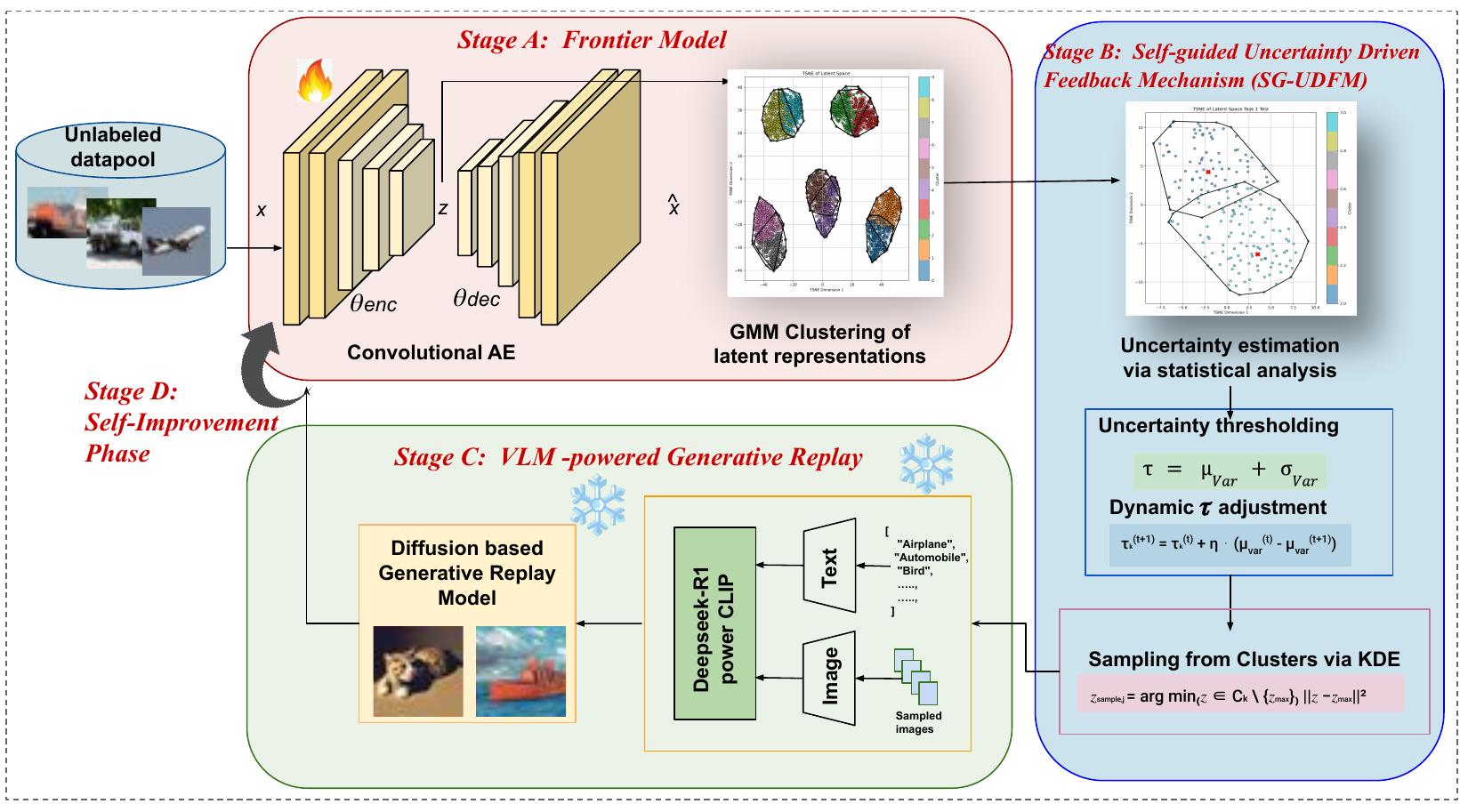}
    \caption{\textcolor{black}{The \textbf{``Replay to Remember (R2R)''} architecture for continual learning with generative replay and uncertainty-driven feedback.}}
    \label{fig:architecture}
    \vspace{-.4cm}
\end{figure*}

This section presents an overview of our proposed ``Replay to Remember (R2R)" framework in continual learning. Referring to Fig.\ref{fig:architecture}, the R2R framework is structured into several stages: (Stage A) \textit{Frontier Model}, where unlabelled data is categorized into clusters representing distinct classes; (Stage B) \textit{Self-Guided Uncertainty-driven Feedback Mechanism}, that assesses the reliability of class clusters; (Stage C) a \textit{VLM-powered Generative Replay} module that generates synthetic samples aimed at reinforcing these weaker class clusters; (Stage D) \textit{Self-improvement Phase} enables continuous adjustment, where the frontier model explores through new data using some pretext tasks, reinforcing learning without catastrophic forgetting.

\vspace{-11pt}
\subsection{Stage A: Frontier Model}

The Frontier Model forms the backbone of our continual learning framework, categorizing unlabelled data into clusters representing distinct classes. This model leverages a Convolutional Autoencoder (CAE) architecture \cite{Kingma2013AutoEncodingVB}, efficiently encoding input data into a compact latent space representation. The CAE captures hierarchical spatial features and patterns, enabling it to cluster unlabelled data based on structural similarities. By dynamically analyzing latent embeddings, the model identifies emerging patterns and adapts its clustering strategy, ensuring scalability and resource-efficient learning without compromising accuracy.

\vspace{-9pt}
\subsubsection{Continual Network Adaptation}  
\textcolor{black}{The Frontier Model employs a Convolutional Autoencoder (CAE) to extract compact latent representations from unlabelled data, enabling unsupervised clustering and continual adaptation. The encoder compresses each input $\mathbf{x}$ into a latent vector $\mathbf{z}$, while the decoder reconstructs the input from this representation:}
\vspace{-.2cm}
\begin{equation}
\mathbf{z} = f_\text{encoder}(\mathbf{x}; \theta_\text{enc}), \quad \hat{\mathbf{x}} = f_\text{decoder}(\mathbf{z}; \theta_\text{dec}),
\end{equation}
where $\theta_\text{enc}$ and $\theta_\text{dec}$ denote the parameters of the encoder and decoder, respectively. The model is trained to minimize the reconstruction loss, ensuring that the latent vector $\mathbf{z}$ captures spatial and structural features relevant for downstream clustering:
\begin{equation}
\mathcal{L}_\text{recon} = \| \mathbf{x} - \hat{\mathbf{x}} \|^2_2,
\end{equation}
\textcolor{black}{The CAE is preferred due to its lightweight architecture, ability to preserve local structure, and compatibility with unsupervised settings, making it ideal for representation learning without requiring labels or pretraining. To mitigate feature shift across tasks \textcolor{black}{ \cite{Wang2023ACS}}, an issue commonly observed in continual learning, we introduce an uncertainty-aware replay mechanism (detailed in Section 3.2). When high-variance clusters are detected in latent space, semantically guided synthetic samples are generated and replayed to reinforce their structure. This enables the CAE to retain useful representations without requiring storage of prior task data.}

\subsubsection{Clustering Using Gaussian Mixture Model (GMM)}

As shown in Fig.\ref{fig:architecture}, the framework employs the Gaussian Mixture Model (GMM) to effectively group samples based on their latent representations. The steps of the clustering procedure are outlined below: Given an input sample \( x_i \in \mathcal{X} \), the encoder of the Convolutional Autoencoder (CAE) maps the sample to a latent representation \( z_i \):
\begin{equation}
z_i = f_{\text{enc}}(x_i), \quad z_i \in \mathbb{R}^d,
\end{equation}
where \( f_{\text{enc}} \) represents the encoder function, and \( d \) is the dimension of the latent space. These latent vectors \( \{z_1, z_2, \dots, z_N\} \) for \( N \) samples are compact representations that encapsulate the salient features of the input data.
The latent vectors are provided to a Gaussian Mixture Model (GMM), which assumes that the data in the latent space is generated from a mixture of \( K \) Gaussian distributions. The GMM models the probability density function as:
\begin{equation}
p(z_i) = \sum_{k=1}^K \pi_k \mathcal{N}(z_i \mid \mu_k, \Sigma_k),
\end{equation}
where \( \pi_k \) represents the weight of the \( k \)-th Gaussian component, satisfying the constraint \( \sum_{k=1}^K \pi_k = 1 \). The mean vector of the \( k \)-th Gaussian is denoted by \( \mu_k \), while \( \Sigma_k \) corresponds to its covariance matrix. The Gaussian probability density function (PDF) for the \( k \)-th component, \( \mathcal{N}(z_i \mid \mu_k, \Sigma_k) \), is expressed as:
\begin{equation}
\mathcal{N}(z_i \mid \mu_k, \Sigma_k) = \frac{\exp\left(-\frac{1}{2}(z_i - \mu_k)^\top \Sigma_k^{-1} (z_i - \mu_k)\right)}{\sqrt{(2\pi)^d \lvert \Sigma_k \rvert}},
\end{equation}
where \( d \) denotes the dimensionality of the latent representation \( z_i \), and \( \lvert \Sigma_k \rvert \) is the determinant of the covariance matrix. This formulation ensures that the Gaussian mixture effectively captures the structure of the latent space by modeling the data distribution using multiple Gaussian components.
Further, the posterior probability that a sample \( z_i \) belongs to the \( k \)-th Gaussian component is calculated as:
\begin{equation}
\gamma_{ik} = \frac{\pi_k \mathcal{N}(z_i \mid \mu_k, \Sigma_k)}{\sum_{j=1}^K \pi_j \mathcal{N}(z_i \mid \mu_j, \Sigma_j)},
\end{equation}
where \( \gamma_{ik} \) is the responsibility of the \( k \)-th component for \( z_i \). The final cluster assignment for each latent vector is determined as:
\begin{equation}
\text{Cluster}(z_i) = \arg\max_k \gamma_{ik}.
\end{equation}
\vspace{-11pt}

Regarding the iterative refinement of the tasks, the samples \( \{x_1, x_2, \dots, x_N\} \) are passed through the encoder, generating latent vectors \( \{z_1, z_2, \dots, z_N\} \) for each task. The GMM clusters these vectors into \( K \) clusters, ensuring task-specific separation. As new tasks are introduced, the GMM dynamically adapts to handle additional latent representations, effectively clustering data across tasks.

\vspace{0.3em}
\noindent\textbf{Clarification on Clustering Scope:} During training on a new task 
$\mathcal{T}_t$, GMM clustering is performed over the latent representations of both the current task's real data and synthetic samples generated via replay from previously learned tasks. While raw data from earlier tasks is not stored, our framework utilizes uncertainty-aware generative replay (Section~\ref{sec:VLM}) to regenerate approximate distributions of past task data. These generated samples are projected into the same latent space by the CAE and included in the clustering pipeline. This strategy allows the GMM to model the cumulative structure of all tasks up to $\mathcal{T}_t$, preserving representation continuity and facilitating stable cluster assignment, without violating memory constraints inherent to the continual learning setting.

\vspace{-11pt}
\subsection{Stage B: Self-Guided Uncertainty-Driven Feedback Mechanism}
\label{sec:SGUDFM}
Traditional continual learning methods like Experience Replay (ER) \cite{Chaudhry2019ContinualLW} and Learning without Forgetting (LwF) \cite{Li2016LearningWF} apply replay uniformly, without assessing which representations are fragile. Our method addresses this by modelling \textbf{cluster-level uncertainty} via intra-cluster variance, enabling targeted replay of only the most unstable regions in latent space. This improves efficiency and better preserves past knowledge. To address this, we propose a novel \textbf{Self-Guided Uncertainty-driven Feedback Mechanism (SG-UDFM)}, which dynamically identifies uncertain regions and enhances cluster representations through targeted fine-tuning. This mechanism ensures that the model adapts effectively to evolving data distributions, mitigating forgetting and improving robustness in a continual learning setting.

This mechanism is crucial for improving the learning process. It allows the model to focus adaptively on these uncertain regions, ensuring that it retains high representational capacity and effectively learns from evolving data distributions. By dynamically directing attention to areas of uncertainty, the SG-UDFM enhances the model's ability to handle challenging tasks that require robust, incremental learning. The SG-UDFM operates through a series of steps to ensure robust cluster representation and continuous learning, particularly in regions where the model demonstrates high uncertainty.

\subsubsection{Clustering-based Uncertainty Estimation and Thresholding}

To assess uncertainty, we approximate the sample's dispersion from its assigned cluster center using the normalized squared \textit{L2} distance, serving as a proxy for variance:
\vspace{-.3cm}
\begin{equation}
\text{Dispersion}(\mathbf{z}_i) = \frac{1}{d} \sum_{j=1}^{d} (\mathbf{z}_{i,j} - \mathbf{z}_k^{\text{center}}[j])^2,
\end{equation}
where \(\mathbf{z}_{i,j}\) represents the \(j\)-th component of the latent vector, and \(\mathbf{z}_k^{\text{center}}[j]\) is the corresponding component of the cluster center. While this measure is not the statistical variance in the formal sense (which is defined over random variables), it effectively captures intra-cluster spread and is commonly used as a surrogate metric in clustering-based uncertainty estimation. An uncertainty threshold \(\tau_k\) is calculated per cluster based on the empirical mean dispersion (\(\mu_{\text{disp}}^k\)) and standard deviation (\(\sigma_{\text{disp}}^k\)) of the sample dispersions within that cluster:
\vspace{-.2cm}
\begin{equation}
\tau_k = \mu_{\text{disp}}^k + \sigma_{\text{disp}}^k.
\end{equation}
\vspace{-.1cm}
Samples with dispersion exceeding the threshold (\(\text{Dispersion}(\mathbf{z}_i) > \tau_k\)) are flagged as uncertain. These samples are prioritized for synthetic data generation via generative replay (GR) to improve cluster consistency and refine representations.

\subsubsection{Dynamic Threshold Adjustment}

The uncertainty threshold is dynamically updated to adapt to improvements in model performance. After each training task \(t\), the threshold for cluster \(k\) is updated as:
\vspace{-5pt}
\begin{equation}
\tau_k^{(t+1)} = \tau_k^{(t)} + \eta \cdot \left( \mu_{\text{var}}^{(t)} - \mu_{\text{var}}^{(t+1)} \right),
\label{eq:Dyn_threshold}
\end{equation}
\vspace{-.7cm}

where \(\eta\) is the learning rate for threshold adjustment and \(\mu_{\text{var}}^{(t)}\) and \(\mu_{\text{var}}^{(t+1)}\) represent the mean \& variance before and after learning task \(t\). This adaptive update mechanism ensures the model maintains sensitivity to evolving representations, emphasizing regions with high uncertainty to achieve robust feature learning.

\subsubsection{Sampling from Clusters}

Once a cluster is flagged as uncertain, our R2R architecture employs a robust sampling strategy to extract representative samples from the cluster for further processing. This sampling mechanism is underpinned by the use of Gaussian Kernel Density Estimation (KDE) and the identification of the highest density point within the cluster. The process is outlined as follows:

Let \(\mathcal{C}_k = \{\mathbf{z}_1, \mathbf{z}_2, \dots, \mathbf{z}_n\}\) represent the set of latent vectors in the flagged cluster \(k\), where \(\mathbf{z}_i \in \mathbb{R}^d\) denotes the \(i\)-th latent vector in a \(d\)-dimensional latent space. The goal is to identify the most representative samples that encapsulate the characteristics of the cluster while prioritizing high-density regions. To achieve this, we estimate the probability density threshold function of the latent space within the cluster using KDE:

\begin{equation}
f_k(\mathbf{z}) = \frac{1}{n} \sum_{i=1}^n K_h(\mathbf{z} - \mathbf{z}_i),
\label{eq:KDE}
\end{equation}

where \(K_h(\cdot)\) is the kernel function with bandwidth \(h\), which controls the smoothness of the density estimation. We use a Gaussian kernel for \(K_h\), defined as:

\begin{equation}
K_h(\mathbf{z}) = \frac{1}{(2\pi h^2)^{d/2}} \exp\left(-\frac{\|\mathbf{z}\|^2}{2h^2}\right).
\end{equation}

The latent vector \(\mathbf{z}_\text{max}\) corresponding to the highest density point within the cluster is computed as:

\begin{equation}
\mathbf{z}_\text{max} = \arg\max_{\mathbf{z} \in \mathcal{C}_k} f_k(\mathbf{z}).
\end{equation}

This vector \(\mathbf{z}_\text{max}\) represents the mode of the cluster's density distribution and is used as a focal point for subsequent sample extraction. To further augment the representational diversity, additional samples are selected based on their proximity to \(\mathbf{z}_\text{max}\) while maintaining sufficient variation. These samples, denoted as \(\{\mathbf{z}_{\text{sample}, 1}, \mathbf{z}_{\text{sample}, 2}, \dots, \mathbf{z}_{\text{sample}, m}\}\), are chosen using the following criterion:

\begin{equation}
\mathbf{z}_{\text{sample}, j} = \arg\min_{\mathbf{z} \in \mathcal{C}_k \setminus \{\mathbf{z}_\text{max}\}} \|\mathbf{z} - \mathbf{z}_\text{max}\|^2,
\end{equation}

where \(m\) is the desired number of additional samples, and the term ensures that the selected samples are close to the mode while spanning different regions of the cluster.

This sampling methodology ensures that the extracted samples effectively capture the uncertainty and diversity within the flagged cluster, enabling the model to refine its understanding of the latent space and improve class representations. By leveraging KDE and the highest density point, the architecture aligns with best practices in density-based sampling and uncertainty-aware processing.


\subsection{Stage C: VLM-powered Generative Replay}
\label{sec:VLM}
To reinforce learning in uncertain clusters, we use \textbf{Generative Replay (GR)} inspired by neurological memory recall. Synthetic samples are generated using Stable Diffusion v1.4, guided by CLIP-derived class tokens, to reflect real data characteristics (refer Fig.~\ref{fig:architecture}). To improve semantic alignment, we incorporate \textbf{DeepSeek-R1} \cite{DeepSeekAI2025DeepSeekR1IR} into the CLIP prompt pipeline. These samples stabilize cluster boundaries and mitigate forgetting. Continual learning principles are preserved as the model is trained from scratch using only unlabeled data, with pretrained modules (CLIP and DeepSeek-R1) used solely for guiding replay prompts, not for representation learning or classification. The framework iteratively refines latent spaces using both real and replayed data, ensuring robust adaptation across tasks without task IDs, labels, or memory buffers.

\vspace{-11pt}
\subsubsection{VLM-CLIP for synthetic data mapping}

After identifying uncertain clusters, ten representative samples from each cluster are extracted based on the thresholds computed in Eq.(\ref{eq:Dyn_threshold}) and Eq.(\ref{eq:KDE}). To enhance interpretability, we employ DeepSeek R1\footnote{In this work, DeepSeek-R1 is leveraged to generate semantic prompts to guide generative replay, due to its strong reliability with low compute cost. However, advanced GPT-4o or BLIP-2 also could be used as alternatives.}., a state-of-the-art small language model (SLM) \cite{DeepSeekAI2025DeepSeekR1IR}, to generate textual labels These labels are derived from random words semantically similar to class names in well-known benchmark datasets, ensuring consistency with our data distribution.

For visual inputs, ten images from each uncertain cluster are processed using the CLIP model, where the textual embeddings generated by DeepSeek R1 are paired with the visual embeddings of the extracted samples. This mapping allows us to identify the most probable class association for each uncertain cluster based on multimodal 
alignment. Subsequently, these mapped synthetic class names serve as textual prompts for a diffusion model, which generates a labelled synthetic dataset. This synthetic dataset facilitates further processing, enabling robust representation learning while addressing the uncertainty in cluster assignments.

\subsubsection{Dynamic Replay Scheduling with Synthetic Sample Reintegration}

The generative replay process is dynamically scheduled based on the uncertainty threshold \(\tau\) computed during Stage (B). Clusters with uncertainty values exceeding \(\tau\) are prioritized for synthetic sample generation. The threshold \(\tau\) is updated at each training step based on the reduction in class-level uncertainty, computed as shown in Eq.(\ref{eq:Dyn_threshold}), where \(\mu_{\text{var}}^{(t)}\) represents the mean-variance of the latent vectors at time step \(t\), and \(\eta\) is the learning rate for threshold updates. This adaptive mechanism ensures that the replay focuses on areas with persistent uncertainty, gradually refining the model's representation.

The generated samples are labelled according to their cluster mappings, which were established in earlier stages. These synthetic samples are saved in directories named after their corresponding class labels and reintegrated into the training pipeline. Additionally, the mappings between clusters and classes are updated dynamically to maintain consistency with evolving cluster boundaries.

The Generative Replay mechanism thus acts as a continuous feedback loop, enriching the dataset with targeted synthetic samples to address uncertainty and enhance the model’s generalization capacity. By leveraging the CAE’s latent space and decoder, this approach ensures that synthetic data generation remains computationally efficient while preserving feature consistency across tasks.

\vspace{-11pt}

\subsection{Stage D: Self-Improvement Phase}
\label{sec:self_improve}
In our continual learning framework, certain clusters may suffer from suboptimal performance due to noisy data, overlapping features, or high intra-cluster variance. To address this, the \textbf{Cluster-Wise Uncertainty-driven Fine-Tuning} stage identifies underperforming clusters and enhances their representations. These clusters, flagged by an uncertainty threshold \(\tau_{\mathrm{uncertain}}\), are defined as:
\vspace{-5pt}
\begin{equation}
\mathcal{C}_{\mathrm{low}} = \big\{ \mathbf{C}_k \mid \text{Var}(\mathbf{z}_i) > \tau_{\mathrm{uncertain}} \big\},
\end{equation}
\vspace{-.1cm}
where \(\mathbf{C}_k\) represents a cluster, and \(\text{Var}(\mathbf{z}_i)\) denotes the variance of latent embeddings. This targeted fine-tuning improves representation balance, mitigates classification bias, and enhances generalization.

To enhance low-performing clusters, we employ a targeted fine-tuning approach using only synthetic labelled data. Generative Replay (GR) generates labelled synthetic samples exclusively for uncertain clusters, addressing data sparsity and refining representations. The fine-tuning objective for cluster-specific weights \(\mathcal{W}_{\mathrm{cluster}}\) is:
\vspace{-5pt}
\begin{equation}
\mathcal{L}_{\mathrm{fine}} = \lambda_1 \cdot \mathcal{L}_{\mathrm{synthetic}},
\end{equation}
\vspace{-.1cm}
where \(\lambda_1\) governs the contribution of synthetic data. No real data is utilized in the refinement process, ensuring adaptive self-improvement based solely on uncertainty-driven feedback. Cluster-level generalization is enhanced by identifying structurally similar clusters using cosine similarity:
\vspace{-5pt}
\begin{equation}
\mathcal{D}_{\mathrm{similarity}} = \text{cosine}(\mathbf{z}_{\mathrm{mean}}^{\mathbf{C}_a}, \mathbf{z}_{\mathrm{mean}}^{\mathbf{C}_b}),
\end{equation}

where \(\mathbf{z}_{\mathrm{mean}}^{\mathbf{C}_a}\) and \(\mathbf{z}_{\mathrm{mean}}^{\mathbf{C}_b}\) denote the mean latent embeddings of clusters \(\mathbf{C}_a\) and \(\mathbf{C}_b\). If \(\mathcal{D}_{\mathrm{similarity}} > \delta_{\mathrm{sim}}\), fine-tuning is extended across both clusters. Refined clusters are re-evaluated for accuracy and uncertainty reduction, with persistently underperforming clusters flagged for further analysis.  Note that, the evaluation at task \( t \) considers the cumulative test split from tasks \( 1 \) through \( t \), ensuring assessment of both current and past knowledge, mitigating forgetting, and improving generalization.

\vspace{-11pt}
\section{Experimental Setup}

\textbf{Datasets:} We evaluate our Replay to Remember (R2R) model using 5 different widely used datasets such as CIFAR-10, CIFAR-100, The Street View House Numbers (SVHN), CINIC-10 and Tiny ImageNet. CIFAR-10 and CIFAR-100 are widely used image classification datasets, each containing images with a resolution of 32 × 32  pixels. CIFAR-10 consists of 10 classes, while CIFAR-100 comprises 100 classes. Similarly, CINIC-10 and SVHN contain 10 classes with images of 32 × 32 resolution. Tiny ImageNet, a more complex dataset, includes 200 classes, with higher-resolution images of 64 × 64 pixels, making it more challenging for classification tasks. All the datasets are divided into five subsets or tasks, where each task is a disjoint set and no classes are repeated in multiple tasks.

\textbf{Evaluation protocols:}
We evaluate our proposed architecture using the standard image classification and clustering evaluation metrics i.e. classification accuracy reported from across all the tasks (mean) for each dataset and silhouette score as given by eq.(\ref{eq:similarity_score}), to assess the quality of the cluster for known or unknown class cluster as given in eq.(\ref{eq:average_similarity}).
\vspace{-5pt}
\begin{equation}
s(i) = \frac{b(i) - a(i)}{\max(a(i), b(i))}
\label{eq:similarity_score}
\end{equation}

where \( s(i) \) represents the Silhouette score for a single sample \( i \), \( a(i) \) denotes the intra-cluster distance, and \( b(i) \) denotes the inter-cluster distance. The overall silhouette score across N number of samples can be defined as:
\vspace{-6pt}
\begin{equation}
S = \frac{1}{N} \sum_{i=1}^{N} s(i)
\label{eq:average_similarity}
\end{equation}

\textcolor{black}{\noindent\textbf{Training Procedure:} During training on a new task ($\mathcal{T}_t$), GMM clustering is performed over the latent representations of both the current task's real data and synthetic samples generated via replay from previously learned tasks. While raw data from earlier tasks is not stored, our framework utilizes uncertainty-aware generative replay (Section~\ref{sec:VLM}) to regenerate approximate distributions of past task data. These generated samples are projected into the same latent space by the CAE and included in the clustering pipeline. This strategy allows the GMM to model the cumulative structure of all tasks up to $\mathcal{T}_t$, preserving representation continuity and facilitating stable cluster assignment, without violating memory constraints inherent to the continual learning setting.}

\textbf{Implementation details:} All the samples from the mentioned dataset are used in their original form. Our baseline model's performance was evaluated using all the test splits from prior tasks. For example, if the model is being tested at task \textit{t4}, then the testing mechanism is as follows: \textit{t1+t2+t3+t4}. This cumulative testing after every task training highlights knowledge retention and catastrophic forgetting. The network is trained using an Adaptive Moment Estimation (Adam) optimizer with the following hyperparameters: lr: 1e-3 (0.001), Decay Rate (gamma): 0.9, Loss Function: Mean Squared Error (MSE), batch size = 32. For all methods and the upper bound method with the full training data, we train 100 epochs with a Mean Squared Error (MSE)
loss function. The proposed method is implemented using the PyTorch framework. The implementation was done in a machine with NVIDIA DGX A100 GPU with 24GB RAM and takes around 8 hours to train the model.


\vspace{-10pt}

\section{Experimental Results}
Various quantitative and qualitative analyses are carried out in the CIFAR-10, CIFAR-100, CINIC-10, TinyImageNet and SVHN datasets to verify the effectiveness of our proposed framework. 

\begin{table}[t!]
\caption{Without GR vs. With GR model mean test accuracy results with dataset sizes (UL: Unlabeled, Test: Test set sizes).}
\footnotesize
\label{tab:WoutGR_vs_WGR}
\centering
\renewcommand{\arraystretch}{1.1} 
\setlength{\tabcolsep}{5pt} 
\begin{tabular}{|l|cc|cc|}
\toprule
\textbf{Datasets}   & \textbf{UL} & \textbf{Test} & \textbf{Without using GR} & \textbf{With using GR} \\
\midrule
SVHN               & 70k & 26k & 48.56\%                 & \textbf{95.18\%}        \\
CIFAR-10           & 42k & 10k & 33.82\%                  & \textbf{98.13\%}        \\
CIFAR-100           & 30k & 15k  & 12.44\%                 & \textbf{73.06\%}        \\
CINIC-10           & 150k  & 90k & 32.71\%                 & \textbf{93.41\%}        \\
TinyImageNet       & 60k & 30k  & 8.12\%                 & \textbf{59.74\%}        \\
\bottomrule
\end{tabular}
\end{table}

\vspace{-.5cm}
\subsection{Dataset Analysis}
The analysis of various datasets utilized in our experimentation for R2R is depicted in Table\ref{tab:WoutGR_vs_WGR}. Supporting our no-pertaining policy, no original labels are used from the dataset in all the experiments. Analogous to prior works in the Continual Learning domain \cite{design_radhakrishnan_2024}, a similar setting is adapted for data distribution, relying solely on unlabeled samples during training. Specifically, unlabelled data ranging from 30k to 150k as shown in Table.\ref{tab:WoutGR_vs_WGR}. This unsupervised continual learning approach eliminates the need for manual labels, mirroring real-world scenarios where labelled data is scarce. By leveraging the structure of unlabeled data, our method demonstrates the effectiveness of self-guided statistical learning in extracting meaningful representations and achieving robust performance.

\vspace{-15pt}
\subsection{R2R Performance Analysis}

\begin{figure}[h!]
    \centering
\includegraphics[width=\textwidth, height=5cm, keepaspectratio]{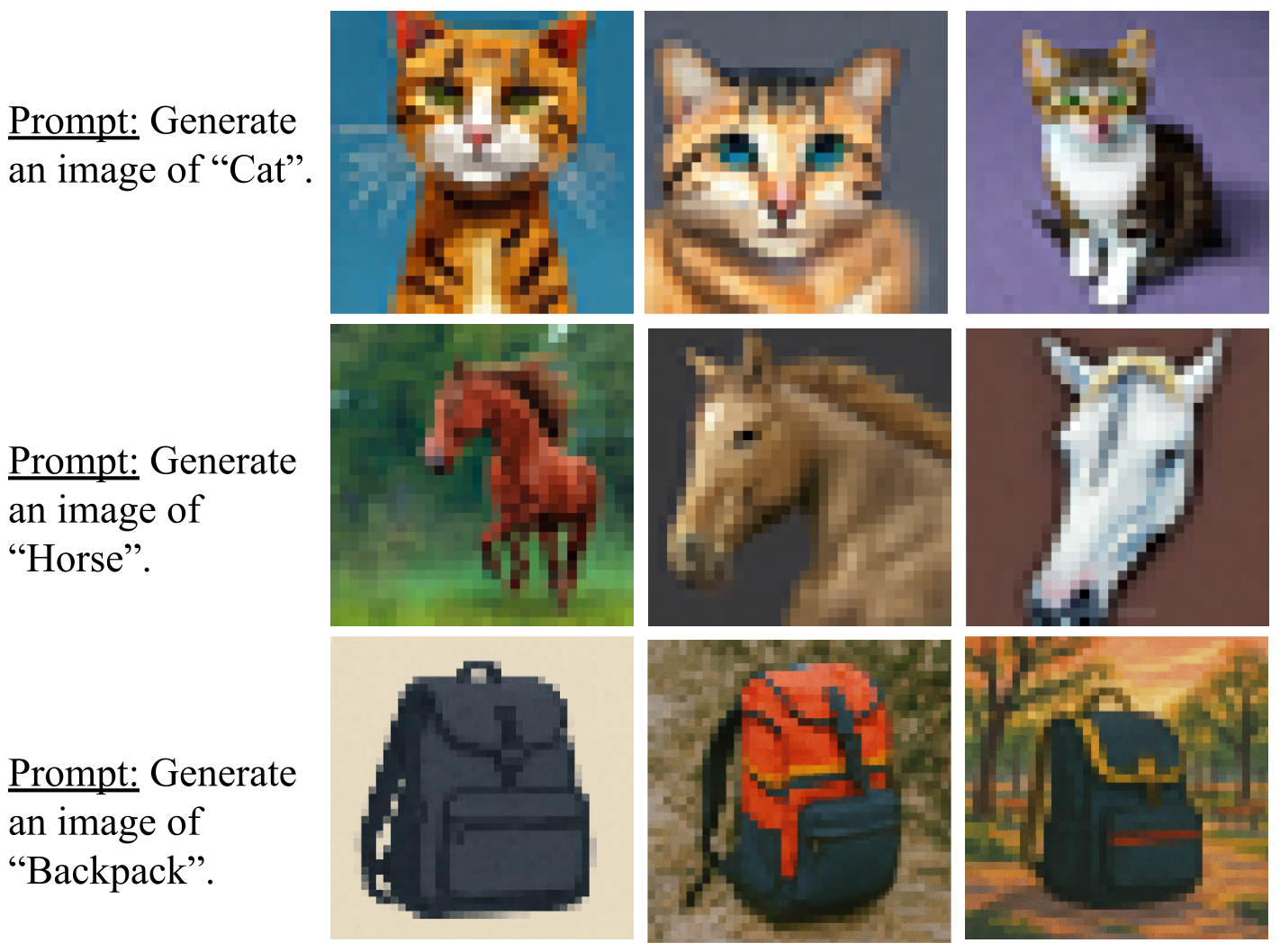} 
    \vspace{-12pt}
    \caption{Illustration of selected Generative Replay (GR) samples from VLM-powered GR stage.}
\label{eq:GR}
\end{figure}

Various experiments are carried out to analyse the performance model's mean accuracies as depicted in Table.~\ref{tab:WoutGR_vs_WGR}, without any initial pertaining. It can be observed that our R2R framework using generative replay achieves 95.18\%, 98.13\%, 73.06\%, 93.41\%, 59.74\% accuracies in SVHN, CIFAR-10, CIFAR-100, CINIC-10 and TinyImageNet, respectively. In comparison with the naive model without using GR a significant increment in accuracies i.e. 46.62\%, 64.31\%, 60.62\%, 60.79\%, and 51.62\% are reported for SVHN-10, CIFAR-10, CIFAR-100, CINIC-10 and TinyImageNet datasets, respectively. 

\textcolor{black}{Furthermore, to visually comprehend the generative replay, some of the GR samples synthetically generated via CLIP-based prompting are visually depicted in Fig.~\ref{eq:GR}, wherein even without any initial pretraining, the labels are made available in a pseudo-synthetic fashion with the help of SG-UDFM module. To prevent feature shifts, the generated GR samples are maintained with the same image dimensions and quality as the original datasets. }


\vspace{-11pt}
\subsection{R2R Task-wise Performance Analysis}
\begin{figure}[t!]
    \centering
    \includegraphics[width=\textwidth, height=4cm, keepaspectratio]{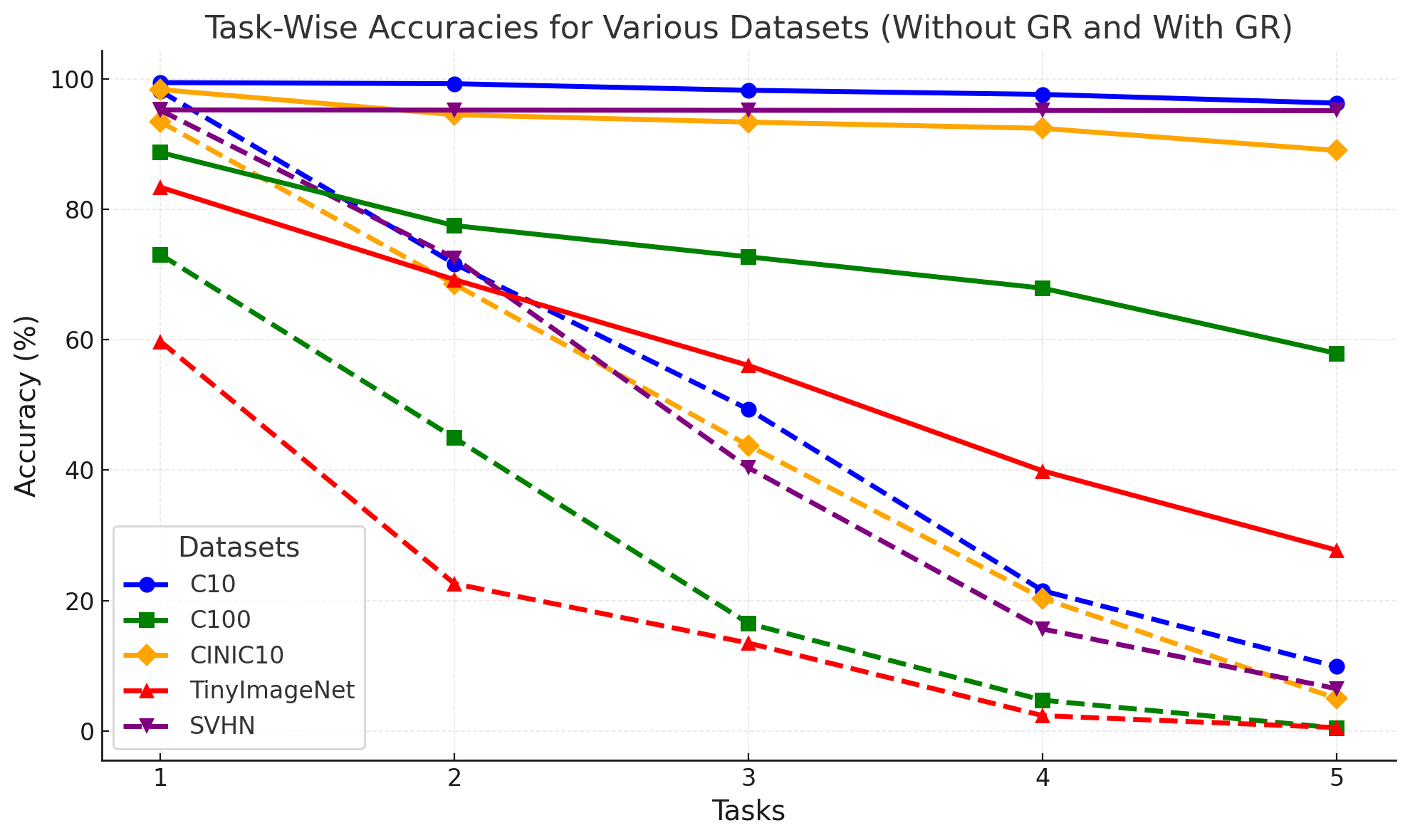} 
    \vspace{-12pt}
    \caption{Illustration of task-wise performances of all the datasets. Continuous lines refer to 'Using GR' and dotted lines refer to 'Without GR'. (Best viewed in colour).  }
\label{eq:taskwise}
\end{figure}

The task-wise performance analysis of the R2R framework is carried out by considering all the cumulative prior tasks' test split. For instance, episodic task 3 evaluation involves the cumulative test split until task 3 i.e. \textit{t1+t2+t3}. Referring to Fig. \ref{eq:taskwise} that illustrates task-wise performance for 5 episodic tasks across multiple datasets, the amount of catastrophic forgetting and knowledge retention can be well analyzed.

It is observed that without generative replay (\textit{dotted lines}), a common trend observed in all the datasets is the degradation of model accuracies due to catastrophic forgetting. Among different datasets, while `using GR' (\textit{continuous lines}), CIFAR 10 (task1-99.43\%, task 5- 96.28\%), CINIC-10 (task1-88.71\%, task 5- 57.89\%) and SVHN-10 (task1 - 95.24\%, task2 - 95.12\%) showed outperformance in terms of consistent accuracy even after 5 episodic tasks. Whereas,  other two datasets i.e. TinyImageNet(task1-83.38\%, task5 -27.71\%) and CIFAR-100 (task1-88.71\%, task 5- 57.89\%) showed slight degradation in performance. This performance trend could be attributed to the higher inter-class variability in those datasets, which may require stronger generative replay strategies to preserve feature representations over multiple tasks.


\vspace{-10pt}
\subsection{Ablation Study}
\subsubsection{Impact of Self-Guided Uncertainty-Driven Feedback Mechanism}

\begin{figure}[t!]
    \centering
\includegraphics[width=\textwidth, height=4cm, keepaspectratio]{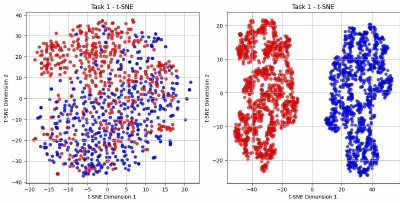} 
    \vspace{-12pt}
    \caption{Illustration of GMM clustering of latent representations of task 1 (2 classes) of CIFAR-10, before and after self-guided uncertainty driven-feedback mechanism.}
\label{eq:before_after}
\end{figure}

The Self-Guided Uncertainty-Driven Feedback Mechanism (SG-UDFM) enhances cluster consistency and representation in unsupervised continual learning by dynamically identifying and refining ambiguous samples via uncertainty-based thresholding and generative replay (GR). The latent representations of GMM are shown in terms of t-SNE \cite{NIPS2002_6150ccc6} in Fig.~\ref{eq:before_after}. As shown in the figure, cluster quality at task 1 before applying the R2R framework was distorted, yielding a silhouette score of 0.14 (left image). Whereas, after the R2R feedback mechanism i.e. SG-UDFM, clusters became more distinct, improving the silhouette score to 0.72 (right image). This demonstrates SG-UDFM’s ability to reduce high-variance samples, reinforce decision boundaries, and maintain a stable latent space, improving clustering accuracy across tasks.

\vspace{-.3cm}
\subsubsection{Impact of our `No Pretrain' Policy}
The `No Pretrain Policy' ensures that the model starts learning without any prior knowledge, relying solely on unsupervised feature extraction. To empirically show its real impact in CL, quantitative analysis of ``No pretaining" vs ``Pretraining" is carried out as shown in Table.\ref{tab:thresholds}. It can be observed that the former achieves competitive results against fully supervised latter approach (e.g. 0.358 vs 0.304 in CIFAR-10, 0.382 VS 0.325 in CIFAR-100) . Nevertheless, the proposed R2R ``No pretaining" approach policy forces the model to learn representations from scratch, making it more relevant and adaptable to continual learning scenarios.

 \vspace{-5pt}

\begin{table}[h]
    \centering
    \caption{Mean threshold values per dataset computed across all tasks. Each value represents the mean ± standard deviation of the threshold averaged across three channels. The "With Pretrained" column shows values by pretraining with ImageNet.}
    \label{tab:thresholds}
    \begin{tabular}{lcc}
        \toprule
        \textbf{Dataset} & \textbf{Without Pretraining} & \textbf{Pretrained} \\
        \midrule
        SVHN        & 0.337 $\pm$ 0.011 & 0.286 $\pm$ 0.009 \\
        CIFAR-10    & 0.358 $\pm$ 0.028 & 0.304 $\pm$ 0.024 \\
        CIFAR-100   & 0.382 $\pm$ 0.029 & 0.325 $\pm$ 0.025 \\
        CINIC-10    & 0.359 $\pm$ 0.026 & 0.301 $\pm$ 0.028 \\
        TinyImageNet & 0.415 $\pm$ 0.062 & 0.353 $\pm$ 0.053 \\
        \bottomrule
    \end{tabular}
\end{table}

\vspace{-.4cm}
\subsubsection{Impact of sampling quantity}
\begin{figure}[t]
    \centering
    \includegraphics[width=\textwidth, height=4.5cm, keepaspectratio]{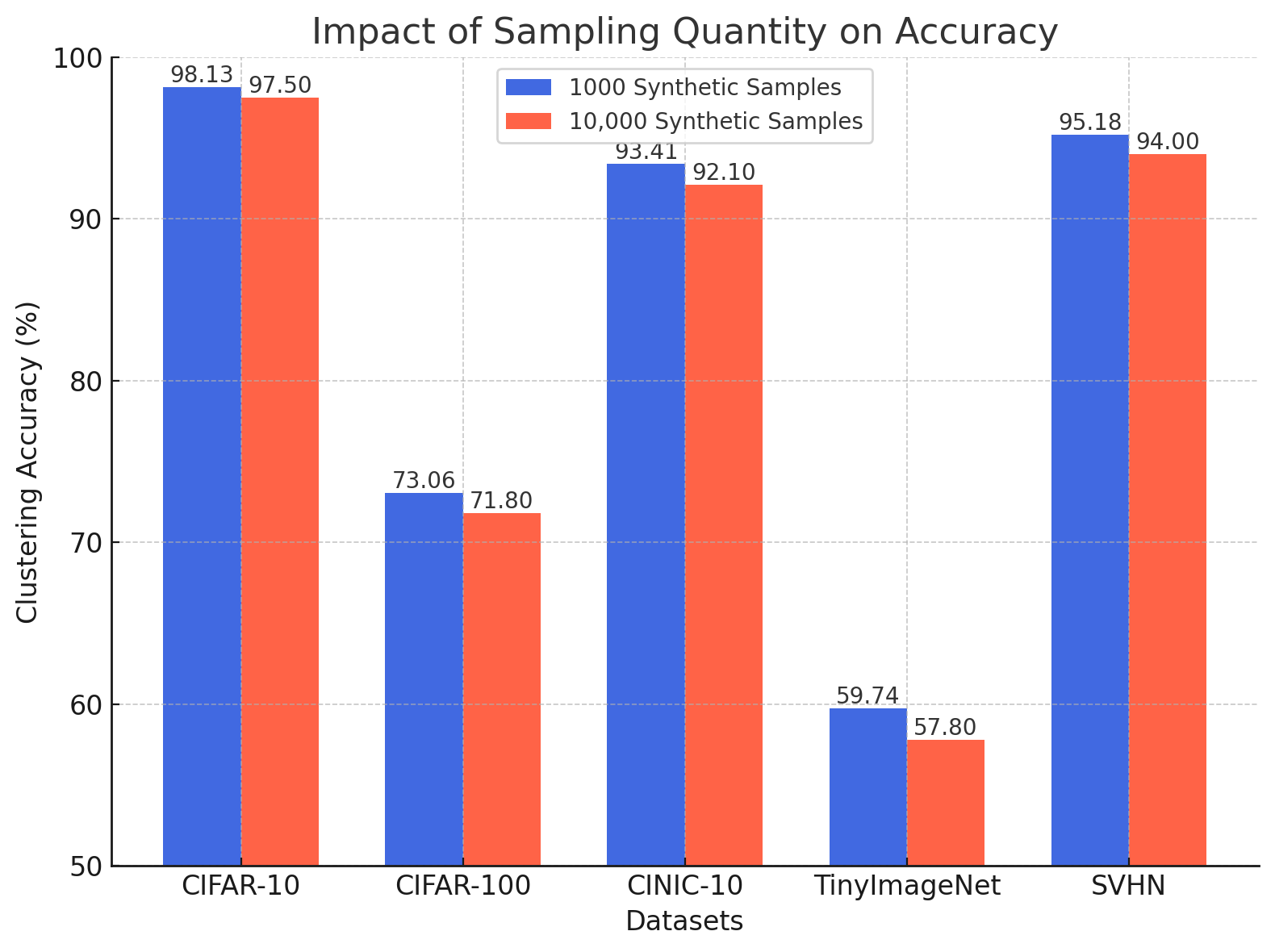} 
    \vspace{-.3cm}
    \caption{Illustration of dataset-wise performances with varying labelled synthetic dataset sample sizes.}
\label{fig:sampling}
\end{figure}
The quantity of sampled synthetic data is crucial in refining feature representations and reducing uncertainty in unsupervised continual learning. As shown in Fig. \ref{fig:sampling}, our observations show that a moderate amount (1k synthetic samples per query class) of synthetic labelled data significantly enhances cluster stability like 98.13\% \& 95.18\% accuracies in CIFAR-10 and SVHN datasets, and reduces uncertain samples by reinforcing weak decision boundaries. However, beyond a certain threshold, adding more synthetic samples provides diminishing returns, as the model has already captured most of the necessary feature variations. For instance, 
for CIFAR-10 and SVHN datasets, with surplus synthetic labelled data of up to 10k samples, the accuracies reduce to 97.5\% and 94\%, respectively.  Excessive synthetic data can also introduce redundant information, slowing down convergence without substantial gains in clustering accuracy.
\vspace{-.6cm}
\subsection{State-of-the-art comparision}
\vspace{-.5cm}
\begin{table}[h!]
    \centering
    \begin{tabular}{l c c c}
        \hline
        \textbf{Dataset} & \textbf{NS } & \textbf{EST} & \textbf{R2R (Ours)}  \\
        \hline
        SVHN & 91.65\% & \textit{93.00\%} & \textbf{95.18\%}  \\
        CIFAR-10 &  89.15\% & \textit{94.21\%} & \textbf{98.13\%}  \\
        CIFAR-100 & 70.53\% & \textbf{76.42\%} & \textit{73.06\%}  \\
        CINIC-10 & 83.47\% & \textit{88.59\%} & \textbf{93.41\%} \\
        TinyImageNet & 49.32\% & \textit{52.23\%} & \textbf{59.74\%} \\
        \hline
    \end{tabular}
    \caption{Comparison of accuracies among NS, EST and our R2R approach. The best two accuracies for each dataset are highlighted in \textbf{bold} and \textit{italics} respectively.}
    \label{tab:accuracy_comparison}
\end{table}
\vspace{-.5cm}
Our method is compared against the recent state-of-the-art approaches like Noisy Student (NS) \cite{9156610} and Enhanced self-training (EST) \cite{design_radhakrishnan_2024}, which are semi-supervised learning method that suffers from confirmation bias. Unlike NS and EST, our R2R framework dynamically refines latent space representations using adaptive uncertainty quantification. Referring to the comparison table as shown in Table. \ref{tab:accuracy_comparison}, \textcolor{black}{it can be observed that R2R framework significantly outperforms the state-of-the-art approaches NS \cite{9156610} and EST \cite{design_radhakrishnan_2024}, with a mean accuracy improvement of +7.08\% and +3.01\%, respectively.}

\vspace{-11pt}
\section{Conclusions and Future work}
We proposed a novel unsupervised continual learning approach, i.e. \textbf{``Replay to Remember(R2R)'' framework}. R2R eliminates the need to use pre-trained weights in the initial setup, as is commonly seen in pseudo-labelling, semi-supervised, and self-supervised methods. We demonstrated that the integration of Self-Guided Uncertainty-Driven Feedback Mechanism (SG-UDFM) and VLM-powered Generative Replay modules into the pipeline reduced the catastrophic forgetting issue, thereby maintaining a good plasticity and stability tradeoff, achieving an average of 4.36\% accuracy against the recent state-of-the-art works. Our framework tackles key challenges in continual learning, including efficient use of labelled knowledge and context awareness, allowing the system to stop learning new tasks when needed. This makes it scalable for real-world applications like autonomous systems and adaptive decision-making in dynamic environments. In future work, we envisage incorporating open-set recognition and using contrastive learning to enhance feature representation for distinguishing known and unknown classes.

\bibliography{mybibfile}

\end{document}